%% file: keystone-arxiv.tex
\documentclass[10pt,letterpaper,twocolumn]{article}
\pdfoutput=1
\usepackage{times}
\usepackage{graphicx}
\usepackage{balance}  
\usepackage{amsmath}
\usepackage{amssymb}
\usepackage{listings}
\usepackage{framed}
\usepackage{float}
\usepackage{multirow}
\usepackage{url}
\usepackage{times}
\usepackage{courier}
\usepackage{graphicx}
\usepackage{stfloats}
\usepackage{color}
\usepackage{caption}
\usepackage{subcaption}
\usepackage{diagbox}
\usepackage{xspace}
\usepackage{epigraph}
\usepackage{booktabs}
\usepackage{tablefootnote}
\usepackage[table,xcdraw]{xcolor}
\usepackage{highlighter}
\usepackage[linesnumbered]{algorithm2e}
\usepackage{listings}
\newcommand{\ra}[1]{\renewcommand{\arraystretch}{#1}}
\SetCommentSty{mycommfont}

\newenvironment{myitemize}
{
   \vspace{0mm}
    \begin{list}{$\bullet$ }{}
        \setlength{\topsep}{0em}
        \setlength{\parskip}{0pt}
        \setlength{\partopsep}{0pt}
        \setlength{\parsep}{0pt}
        \setlength{\itemsep}{1mm}
}
{
    \end{list}
}

\lstdefinelanguage{scala}{
  morekeywords={abstract,case,catch,class,def,%
    do,else,extends,false,final,finally,%
    for,if,implicit,import,match,mixin,%
    new,null,object,override,package,%
    private,protected,requires,return,sealed,%
    super,this,throw,trait,true,try,%
    type,val,var,while,with,yield},
  otherkeywords={=>,<-,<\%,<:,>:,\#,@},
  sensitive=true,
  morecomment=[l]{//},
  morecomment=[n]{/*}{*/},
  morestring=[b]",
  morestring=[b]',
  morestring=[b]"""
}

\def\keystone{KeystoneML}
\def\reals{\mathbb{R}}

\begin{document}

\title{KeystoneML: Optimizing Pipelines for Large-Scale\\ Advanced Analytics}
\date{}
\author{
Evan R. Sparks, Shivaram Venkataraman, Tomer Kaftan, Michael Franklin, Benjamin Recht\\
AMPLab, University  of California, Berkeley,\\
\{sparks,shivaram,tomerk11,franklin,brecht\}@cs.berkeley.edu
}

\maketitle
\begin{abstract}
Modern advanced analytics applications make use of machine learning techniques and contain multiple steps of domain-specific and general-purpose processing with high resource requirements. We present {\keystone}, a system that captures and optimizes the end-to-end large-scale machine learning applications for high-throughput training in a distributed environment with a high-level API. This approach offers increased ease of use and higher performance over existing systems for large scale learning. We demonstrate the effectiveness of {\keystone} in achieving high quality statistical accuracy and scalable training using real world datasets in several domains.
By optimizing execution {\keystone} achieves up to $15\times$ training throughput over unoptimized execution on a real image classification application.
\end{abstract}

\input{introduction_new}
\input{pipeline-background}
\input{nodeopt}
\input{pipelineopt}
\input{evaluation}
\input{related}
\input{conclusion}

\noindent\textbf{Acknowledgements}: We would like to thank Xiangrui Meng, Joseph Bradley for their help in design discussions and Henry Milner, Daniel Brucker, Gylfi Gudmundsson, Zongheng Yang, Vaishaal Shankar for their contributions to the {\keystone} source code. We would also like to thank Peter Alvaro, Peter Bailis, Joseph Gonzales, Nick Lanham, Aurojit Panda, Ameet Talwarkar for their feedback on earlier versions of this paper. This research is supported in part by NSF CISE Expeditions Award CCF-1139158, DOE Award SN10040 DE-SC0012463, and DARPA XData Award FA8750-12-2-0331, and gifts from Amazon Web Services, Google, IBM, SAP, The Thomas and Stacey Siebel Foundation, Adatao, Adobe, Apple, Inc., Blue Goji, Bosch, Cisco, Cray, Cloudera, EMC2, Ericsson, Facebook, Guavus, HP, Huawei, Informatica, Intel, Microsoft, NetApp, Pivotal, Samsung, Schlumberger, Splunk, Virdata and VMware.

\balance

{\small
 \bibliographystyle{abbrv}
 \bibliography{refs}
}
\end{document}

%% file: introduction_new.tex
\section{Introduction}
\label{sec:introduction}

Today's advanced analytics applications increasingly use machine learning (ML) as a core technique
in areas ranging from business intelligence to recommendation to natural language
processing~\cite{manning2003optimization} and speech recognition~\cite{huang2014kernel}.
Practitioners build complex, multi-stage \emph{pipelines} involving feature extraction, dimensionality reduction, data transformations, and training supervised learning models
to achieve high accuracy~\cite{sanchez2013image}.
However, current systems provide little support for automating the construction and optimization of these pipelines.

To assemble such pipelines, developers typically piece together domain specific libraries\footnote{e.g. OpenCV for Images (http://opencv.org/), Kaldi for Speech (http://kaldi-asr.org/)} for feature extraction and general purpose numerical optimization packages~\cite{langford2007vowpal,mllib} for supervised learning.
This is often a cumbersome and error-prone process~\cite{sculley2014}.
Further, these pipelines need to be completely re-engineered when the training data or features grow by an order of magnitude--often the difference between an
application that provides good statistical accuracy and one that does
not~\cite{halevy2009unreasonable}. As no broader system has purview of the end-to-end
application, only narrow optimizations can be applied.

These challenges motivate the need for a system that 
\begin{myitemize}
	\item Allows users to specify end-to-end ML applications in a single system using high level logical operators.
	\item Scales out dynamically as data volumes and problem complexity change.
	\item Automatically optimizes these applications given a library of ML operators and the user's compute resources.
\end{myitemize}

While existing efforts in the data management
community~\cite{hellerstein2012madlib,ghoting2011systemml,mllib} and in the broader machine learning
systems community~\cite{langford2007vowpal,scikit-learn,tensorflow2015-whitepaper} have built
systems to address some of these problems, each of them misses the mark on at least one of the points above. 

\begin{figure}
  \centering
  \includegraphics[width=0.7\columnwidth]{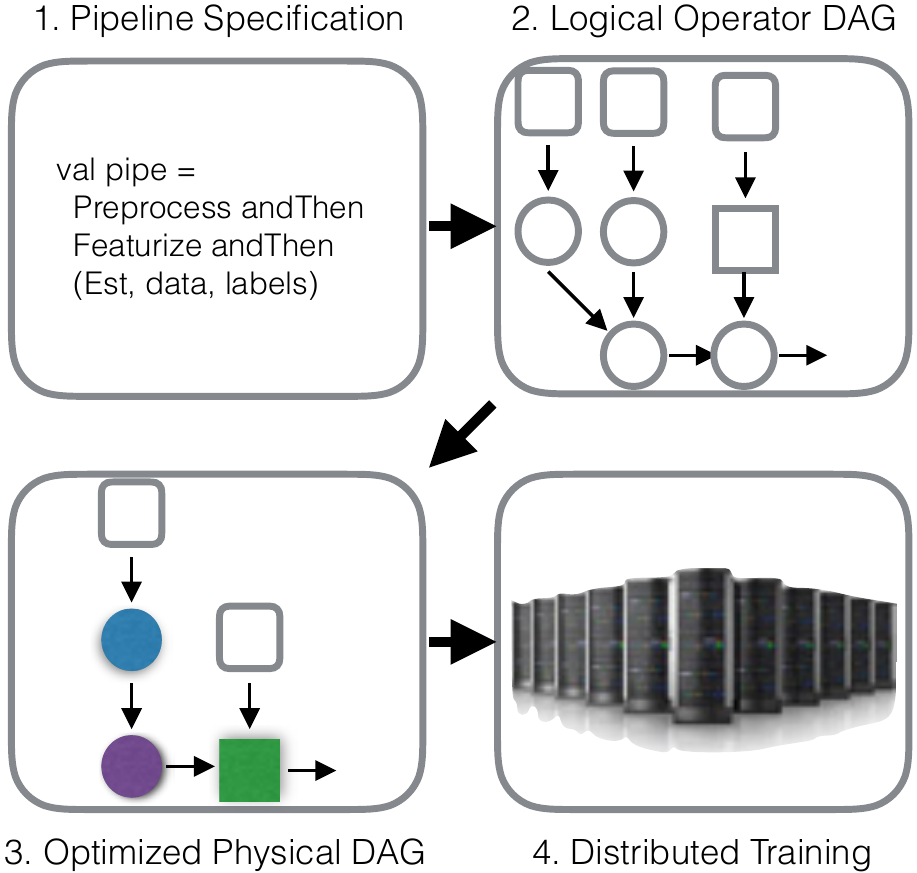}
  \caption{{\keystone} takes a high-level ML application specification, optimizes and trains it in a distributed environment. The trained pipeline is used to make predictions on new data.}
  \label{fig:keystoneflow}
\end{figure}

We present {\keystone}, a framework for ML pipelines designed to satisfy the above requirements. Fundamental to the design of {\keystone} is the observation that model training is only one component of an ML application. While a significant body of recent work has focused on high performance algorithms~\cite{zhang2014dimmwitted,recht2011hogwild}, and scalable implementations~\cite{crotty2014tupleware, mllib} for model training, they do not capture the featurization process or the logical intent of the workflow.
{\keystone} provides a high-level, type-safe API built around \emph{logical operators} to capture end-to-end applications.

To optimize ML pipelines, database query optimization provides a natural motivation for the core design of such a system~\cite{kraska2013mlbase}.
However, compared to relational database query optimization, ML applications present an additional set of concerns. 
First, ML operators are often \emph{iterative} and may require multiple passes over their inputs,
presenting opportunities for data reuse. 
Second, many ML operators provide only approximate answers to their inputs~\cite{recht2011hogwild}.
Third, numerical data properties such as sparsity and dimensionality are a necessary source of information when selecting optimal execution plans and conventional optimizers do not consider such measures.
Finally, the system should be aware of the computation-vs-communication tradeoffs inherent in distributed processing of ML workloads~\cite{ghoting2011systemml,langford2007vowpal} and choose appropriate execution strategies in this regime. 

To address these challenges we develop techniques to do both per-operator optimization and
end-to-end pipeline optimization for ML pipelines. We use a cost-based optimizer that accounts for both computation and 
communication costs and our cost model can easily accommodate new operators and hardware
configurations. To determine which intermediate states are materialized in memory during iterative
execution, we formulate an optimization problem and present a greedy algorithm that works
efficiently and accurately in practice. 

We measure the importance of cost-based optimization and its associated overheads using real-world workloads from computer vision, speech and natural language processing. We find that end-to-end optimization can improve performance by $7\times$ and that physical operator optimizations combined with end-to-end optimizations can improve performance by up to $15\times$ versus unoptimized execution. 
We show that in our experiments, poor physical operator selection can result in up to a $260\times$ slowdown.
Using an image classification pipeline on over 1M images~\cite{sanchez2013image}, we show that {\keystone} provides linear performance scalability across various cluster sizes, and statistical performance comparable to recent results~\cite{Chatfield11,sanchez2013image}. 
Additionally, {\keystone} can match the performance of a specialized phoneme classification system
on a BlueGene supercomputer while using $8\times$ fewer resources. 
In summary, we make the following contributions:
\begin{myitemize}
  \item We present {\keystone}, a system for describing ML applications using high level logical operators. {\keystone} enables end-to-end optimization of ML applications at both the operator and pipeline level.
  \item We demonstrate the importance of physical operator selection in the context of input characteristics of three commonly used logical ML operators, and propose a cost model for making this selection.
  \item We present and evaluate an initial set of whole-pipeline optimizations, including a novel algorithm that automatically identifies a subset of intermediate data to materialize to speed up pipeline execution.
  \item We evaluate these optimizations in the context of real-world pipelines in a diverse set of domains: phoneme classification, image classification, and textual sentiment analysis, and demonstrate near-linear scalability over 100s of machines with strong statistical performance.
  \item We compare {\keystone} with several recent systems for large-scale learning and demonstrate superior runtime from our optimization techniques and scale-out strategy.
\end{myitemize}

{\keystone} is open source software\footnote{\url{http://www.keystone-ml.org/}} and is being used in scientific applications in solar physics~\cite{jonas2016solar} and genomics~\cite{genomicschallenge}

%% file: pipeline-background.tex
\section{Pipeline Construction and Core APIs}
\label{sec:background}
In this section we introduce the {\keystone} API that can be used to express end-to-end ML pipelines. Each pipeline
is composed a number of \emph{operators} that are chained together. For example, Figure~\ref{lst:amazonlisting} 
shows the {\keystone} source code for a complete text classification pipeline. We next describe the building blocks of our API.

\begin{figure}
  \small
\centering
\begin{lstlisting}[basicstyle=\ttfamily\scriptsize, language=Scala, firstline=3, lastline=11]
    val textClassifier = Trim andThen
        LowerCase andThen
        Tokenizer andThen
        NGramsFeaturizer(1 to 2) andThen
        TermFrequency(x => 1) andThen
        (CommonSparseFeatures(1e5), data) andThen
        (LinearSolver(), data, labels)
    val predictions = textClassifier(testData)
\end{lstlisting}
\caption{A text classification pipeline is specified using a small set of logical operators.}
\label{lst:amazonlisting}
\vspace{-0.2in}
\end{figure}

\subsection{Logical ML Operators}
Conventional analytics queries are typically composed using a small number of well studied relational
database operators. This well-defined environment enables important optimizations. However, ML applications lack such an
abstraction and practitioners typically piece together imperative libraries.
Recent efforts have proposed using linear algebra operators such as matrix
multiplication~\cite{ghoting2011systemml}, convex optimization routines~\cite{feng2012towards} or 
multi-dimensional arrays as logical building blocks~\cite{stonebraker2009requirements}. 

In contrast, with {\keystone} we propose a design where high-level ML operations (such as \texttt{PCA},
\texttt{LinearSolver}) are used as building blocks. Our approach has two major benefits: 
First, it simplifies building applications. Even complex pipelines can be built
using just a handful of operators. Second, this higher level abstraction allows us to
perform a wider range of optimizations. Our key insight here is that there are usually multiple well 
studied algorithms for a given ML operator, but that their performance and statistical characteristics 
vary based on the inputs and system configuration. We next describe the API for operators in {\keystone}. 

\begin{figure}
  \small
\centering
\begin{lstlisting}[basicstyle=\ttfamily\scriptsize, language=Scala, firstline=1, lastline=7]
trait Transformer[A, B] extends Pipeline[A, B] {
  def apply(in: Dataset[A]): Dataset[B] =
    in.map(apply)
  def apply(in: A): B
}
\end{lstlisting}
\vspace{-0.1in}
\begin{lstlisting}[basicstyle=\ttfamily\scriptsize, language=Scala, firstline=1, lastline=3]
trait Estimator[A, B] {
  def fit(data: Dataset[A]): Transformer[A, B]
}
\end{lstlisting}
\vspace{-0.1in}
\begin{lstlisting}[basicstyle=\ttfamily\scriptsize, language=Scala, firstline=1, lastline=15]
trait Optimizable[T, A, B] {
  val options: List[(CostModel, T[A,B])]
  def optimize(sample: Dataset[A], d: ResourceDesc): T[A,B]
}

class CostProfile(
  flops: Long, bytes: Long, network: Long)

trait CostModel {
    def cost(sample: Dataset[A], workers: Int): CostProfile
}

trait Iterative {
  def weight: Int
}
\end{lstlisting}
\caption{The {\keystone} API consists of two extendable operator types and interfaces for optimization.}
\label{fig:apioutline}
\vspace{-0.2in}
\end{figure}

Pipelines are composed of \emph{operators}.
Transformers and Estimators are two abstract types of operators in {\keystone}. 
An operator is a function which operates on zero or more inputs to produce some output. 
A \emph{logical operator} satisfies some logical contract. For example, it takes an image and converts it to grayscale.
Every logical operator must have at least one \emph{physical operator} associated with it which implements its logic.
Logical operators with multiple physical implementations are candidates for \emph{optimization}. They are marked \texttt{Optimizable} and have a set of \texttt{CostModels} associated with them.
Operators that are iterative with respect to their inputs are marked \texttt{Iterative}. 

A \emph{Transformer} is an operator that can be applied to individual data items (or to a collection of items) and produces a new data item (or a  collection of data items)--it is a deterministic unary function without side-effects.
Examples of Transformers in {\keystone} include basic data transformations, feature extractors and model application.
The deterministic and side-effect free properties affords the ability to reorder and optimize the execution of the functions without changing the result.

An \emph{Estimator} is applied to a distributed collection of data items and produces a Transformer--it is a function generating function. ML algorithms provided by {\keystone} are Estimators, while featurizers are Transformers. 
For example, \texttt{LinearSolver} is an Estimator that takes a data set and labels, finds the linear model which minimizes the square loss between the training data and labels, and produces a Transformer that can apply this model to new data.

\subsection{Pipeline Construction}
Transformers and Estimators are \emph{chained} together into a \texttt{Pipeline} using a consistent set of rules.
The chaining methods are summarized in Figure~\ref{fig:chainoutline}. 
In addition to linear chaining of nodes using \texttt{andThen}, {\keystone}'s API allows for pipeline branching.
When a developer calls \texttt{andThen} a new \texttt{Pipeline} object is returned. By calling \texttt{andThen} multiple times on the same pipeline, users can create multiple pipelines that branch out. Developers join the output of multiple pipelines of using \texttt{gather}.
Redundancy is eliminated via common sub-expression optimization detailed in Section~\ref{sec:pipelineopt}.
We find these APIs are sufficient for a number of ML applications (Section~\ref{sec:applications}), but expect to extend them over time.

\begin{figure}[t]
\centering
\begin{lstlisting}[basicstyle=\ttfamily\scriptsize, language=Scala, firstline=1, lastline=23]
  trait Pipeline[A,B] {
    def andThen[C](
      next: Pipeline[B, C]): Pipeline[A, C]
    def andThen[C](
      est: Estimator[B, C],
      data: Dataset[A]): Pipeline[A, C]

    // Combine the outputs of branches 
    // into a sequence
    def gather[A, B](branches: Seq[Pipeline[A, B]]): Pipeline[A, Seq[B]]
  }
\end{lstlisting}
\caption{Transformers and Estimators are chained using a syntax designed to allow developers to incrementally build pipelines.}
\label{fig:chainoutline}
\vspace{-0.15in}
\end{figure}

\begin{figure}
  \small
\centering
\includegraphics[width=\columnwidth]{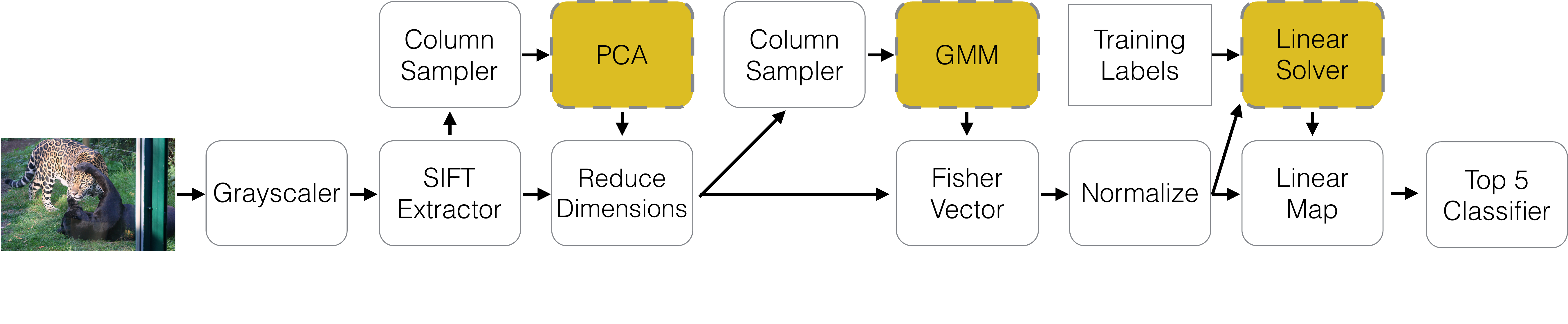}
\caption{A pipeline DAG for image classification. Estimators are shaded.}
\label{fig:imagepipeline}
\vspace{-0.2in}
\end{figure}

\subsection{Pipeline Execution}
\label{sec:pipelineexec}
{\keystone} is designed to run with large, distributed datasets on commodity clusters. Our high level API and optimizers can be executed using any distributed data-flow engine.
The execution flow of {\keystone} is shown in Figure~\ref{fig:keystoneflow}.
First, developers specify pipelines using the {\keystone} APIs described above.
As calls to these APIs are made, {\keystone} incrementally builds an operator DAG for the pipeline.
An example operator DAG for image classification is shown in Figure~\ref{fig:imagepipeline}.
Once a pipeline is applied to some data, this DAG is then optimized using a set of optimizations described below--we call this stage \emph{optimization time}.
Once the application has been optimized, the DAG is traversed depth-first and operators are executed one at a time, with nodes up until pipeline breakers (i.e. Estimators) packed into the same job--this stage is \emph{runtime}.
This lazy optimization procedure gives the optimizer full information about the application in question.
We now consider the optimizations made by {\keystone}.

%% file: nodeopt.tex
\section{Operator-Level Optimization}
\label{sec:nodeopt}

In this section we describe the operator-level optimization procedure used in {\keystone}. Similar to
database query optimizers, the goal of the operator-level optimizer is to choose the best physical
implementation for every machine learning operator in the pipeline. This is challenging to do
because operators in {\keystone} are distributed i.e. they involve computation and communication
across the cluster. Operator performance may also depend on statistical properties like sparsity of
input data and level of accuracy desired.  Finally, as discussed in Section~\ref{sec:background},
{\keystone} consists of a set of high-level operators.  The advantage of having high-level operators
is that we can perform more wide-ranging optimizations. But this makes designing an optimizer more
challenging because unlike relational operators or linear algebra~\cite{ghoting2011systemml},
the set of operators in {\keystone} is not closed. We next discuss how we address these challenges.  

\noindent\textbf{Approach:} The approach we take in {\keystone} is to develop a cost-based optimizer that splits the cost model
into two parts: an operator-specific part and a cluster-specific part. The operator-specific part
models the computation and communication time given statistics of the input data and number of workers and the cluster
specific part is used to weigh their relative importance. More formally, the cost estimate 
for each physical operator, $f$ can be expressed as:
\begin{align}
  c(f, A_s, R) = R_{exec} c_{exec}(f, A_s, R_w) + \\ 
                 R_{coord} c_{coord}(f, A_s, R_w)
\end{align}

Where $f$ is the operator in question, $A_s$ contains statistics of a dataset to be used as its
input, and $R$, the \emph{cluster resource descriptor} represents the cluster computing, memory, and
networking resources available. The cluster resource descriptor is collected via configuration data and microbenchmarks. Statistics captured include per-node CPU throughput (in GFLOP/s), disk and memory bandwidth (GB/s),
and network speed (GB/s), as well as information about the number of nodes available.
$A_s$ is determined through a process we will discuss in Section~\ref{sec:pipelineopt}. 
$R_w$ is the number of cluster nodes available.

The functions, $c_{exec}$, and $c_{coord}$ are developer-defined operator-specific functions (defined as part of the operator \texttt{CostModel}) that
describe execution and coordination costs in terms of the longest critical path in the execution graph of the individual operators~\cite{Williams:2009cx}, e.g. the most FLOPS used by a node in the cluster or the amount of data transferred over the most loaded link. Such functions are also used in the analysis of parallel algorithms~\cite{ballard2013avoiding} and are well known for common linear algebra based operators. $R_{exec}$ and $R_{coord}$ are determined by the optimizer from the cluster resource descriptor ($R$) and capture the relative speed
of local and network resources on the cluster. 

\begin{table}[!t]
\footnotesize
\centering
\begin{tabular}{llll}
\\[-1.8ex]\hline
\hline \\[-1.8ex]
Algorithm                         & Compute        & Network       & Memory \\ 
\hline \\[-1.8ex]
Local QR    & $O(nd(d+k))$   & $O(n(d+k))$   & $O(d(n+k))$ \\
Dist. QR    & $O(\frac{nd(d + k)}{w})$ & $O(d(d+k))$ & $O(\frac{nd}{w} + d^2)$ \\
L-BFGS      & $O(\frac{insk}{w})$   & $O(idk)$  & $O(\frac{ns}{w} + dk)$ \\
Block Solve & $O(\frac{ind(b+k)}{w})$   & $O(id(b+k))$ & $O(\frac{nb}{w} + dk)$ \\
\hline \\[-1.8ex]
\end{tabular}
\caption{Resource requirements for linear solvers. $w$ is the number of workers in the cluster, $i$ the number of passes over the dataset. For the sparse solvers $s$ is the the average number of non-zero features per example, and $b$ is the block size for the block solver. Compute and Memory requirements are per-node, while network requirements are in terms of the data sent over the most loaded link.}
\label{tab:solvertradeoffs}
\vspace{-0.15in}
\end{table}

Splitting the cost model in this fashion allows the the optimizer to easily adapt to new hardware (e.g., GPUs or Infiniband
networks) and also for it to work with both existing and future operators. 
Operator developers only need to implement a \texttt{CostModel} and the system accounts for hardware properties.
Finally we note that the cost model we use here is approximate and that the cost $c$ need not be
equal to the actual running time of the operator. Rather, as in conventional query optimizers, the goal of the cost model is 
to \emph{avoid bad decisions}, which a roughly accurate model will do.
At the boundary of two nearly equivalent operators, either should be acceptable in terms of runtime.
We next illustrate the cost functions for three central operators in {\keystone}
and the performance trade-offs that arise from varying input properties. 

\textbf{Linear Solvers} are supervised Estimators that learn a linear map $X$ between an input
dataset $A$ in $\reals^{n \times d}$ to a labels dataset $B$ in $\reals^{n \times k}$ by finding the
$X$ which minimizes the value $||AX-B||_F$. In a multi-class
classification setting, $n$ is the number of examples or data points, $d$ the number of
features and $k$ the number of classes. In the {\keystone} Standard Library we have several
implementations of linear solvers, distributed and local, including
\begin{myitemize}
\item Exact solvers~\cite{demmel2012communication} that compute closed form solutions to the least squares loss and return an $X$ to
extremely high precision.
\item Block solvers that partition the features into a set of blocks and
use second-order Jacobi or Gauss-Seidel~\cite{bertsekas1989parallel} updates to converge to the right solution.
\item Gradient based methods like SGD~\cite{recht2011hogwild} or L-BFGS~\cite{chen2014large} which
perform iterative updates using the gradient and converge to a globally optimal solution.
\end{myitemize}
Table~\ref{tab:solvertradeoffs} summarizes the cost model for each method.
Constants are omitted for readability but are necessary in practice.

To illustrate these cost tradeoffs empirically, we vary the number of features generated by the
featurization stage of two different pipelines and measure the training time and the 
training loss. We compare the methods on a 16 node cluster.

\begin{figure}[!t]
  \centering
  \includegraphics[width=0.97\columnwidth]{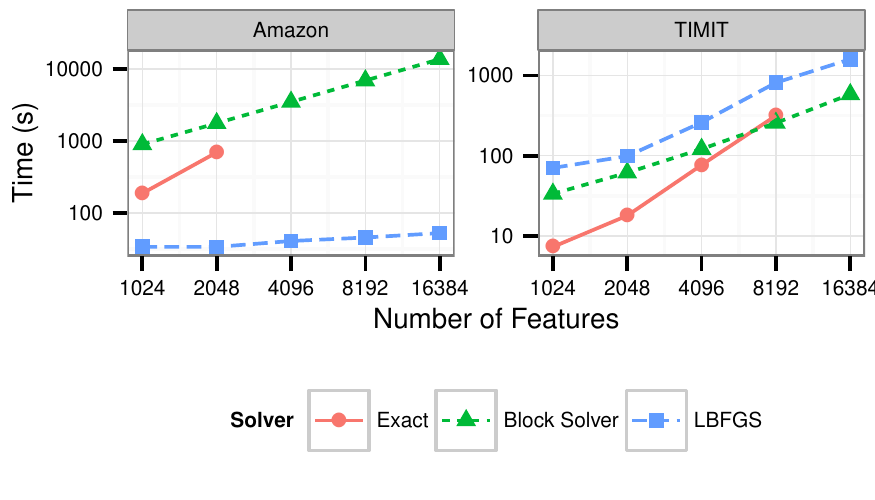}
  \vspace{-0.1in}
  \caption{A poor choice of solver can mean orders of magnitude difference in runtime. Runtime for exact solve grows quadratically in the number of features and times out with 4096 features for Amazon and 16384 features for TIMIT running on 16 \texttt{c3.4xlarge} nodes.}
  \label{fig:solvercomparisontime}
  \vspace{-0.15in}
\end{figure}

On an Amazon Reviews dataset (see Table~\ref{tab:datasets}) with a text classification pipeline, as
we increase the number of features from 1k to 16k we see in Figure~\ref{fig:solvercomparisontime}
that L-BFGS performs $5$-$20\times$ faster than the exact solver and $26$-$260\times$ faster than
the block-wise solver. Additionally the exact solver crashes for greater than 4k features as the
memory requirements are too high. The reason for this speedup is that the features generated in
text classification problems are sparse and the L-BFGS solver exploits the sparse inputs to
calculate gradients cheaply.

The optimal solver choice does not always stay the same as we increase the problem size or as
sparsity changes. For the TIMIT dataset, which has dense features, we see that the exact
solver is $3$-$9\times$ faster than L-BFGS for smaller number of features. However when the number
of features goes beyond 8k we see that the exact solver becomes slower than the block-wise solver
which is also $2$-$3\times$ faster than L-BFGS.  

{\textbf{Principal Component Analysis}} (PCA) is an Estimator used for tasks ranging from
dimensionality reduction to whitening to visualization.  PCA takes an input dataset $A$ in
$\reals^{n \times d}$, and a value $k$ and produces a Transformer which can apply a matrix $P$ in
$\reals^{d \times k}$, where $P$ consists of the first $k$ eigenvectors of the covariance matrix of
$A$.
The $P$ matrix can be found using several techniques including the SVD or via an approximate
algorithm, Truncated SVD~\cite{halko2011}. In our cost model, SVD has runtime 
$O(nd^2)$ and offers an exact answer, while TSVD runs in $O(nk^2)$.
Both methods may parallelized over a cluster.

To better illustrate how the choice of a PCA implementation affects the run time, we construct a
micro-benchmark that varies problem size along $n$, $d$, and $k$, and execute both local and
distributed implementations of the approximate and exact algorithm on a 16-node cluster.  In
Table~\ref{tab:pcacomparison}, we can see that as data volumes increase in $n$ and $d$ it makes
sense to run PCA in a distributed fashion, while for relatively small values of $k$, it can
make sense to use the approximate method.

\begin{table}
\centering
\scriptsize
\ra{1.3}
\begin{tabular}{@{}rrrrcrrr@{}}
\\[-1.8ex]\hline
\hline \\[-1.8ex]
& \multicolumn{3}{c}{$d = 256$} & \phantom{a}& \multicolumn{3}{c}{$d = 4096$}\\
\cmidrule{2-4} \cmidrule{6-8}
& $k=1$ & $16$ & $64$ && $k=16$ & $64$ & $1024$\\ \midrule
$n=10^4$\\
SVD        & \textbf{0.1}  & \textbf{0.1}   & \textbf{0.1}   && 26  & 26    & \textbf{26}    \\
TSVD       & 0.2  & 0.3   & 0.4   && \textbf{3}   & \textbf{6}     & 34    \\
Dist. SVD    & 1.7  & 1.7   & 1.7   && 106 & 106   & 106   \\
Dist. TSVD   & 4.9  & 3.8   & 5.3   && 6   & 22    & 104   \\
$n=10^6$\\
SVD      & 11 & 11  & 11  &&    x   &       x  &        x \\
TSVD     & 14 & 30  & 65  &&    x   &       x  &        x \\
Dist. SVD  & \textbf{2}  & \textbf{2}   & \textbf{2}   && 260 & \textbf{260}   & \textbf{260}   \\
Dist. TSVD & 16 & 59  & 262 && \textbf{75}  & 1,326 & 8,310 \\       
\hline \\[-1.8ex]
\end{tabular}
\caption{Comparison of runtimes (in seconds) for approximate and exact PCA operators across different dataset sizes. A dataset has $n$ examples and $d$ features. $k$ is an algorithm input. An x indicates that the operation did not complete.}
\label{tab:pcacomparison}
\vspace{-0.1in}
\end{table}

{\textbf{Convolution}} is a critical building block of Signal, Speech, and Image Processing
pipelines. In image processing, the Transformer takes in an Image of size $n \times n \times d$ and
applies a bank of $b$ filters (each of size $k \times k$, where $k < n$) to the Image
and returns the $b$ resulting convolved images of size $m \times m$, where $m=n-k+1$. There are three 
main ways to implement convolutions: via a matrix-vector product scheme when convolutions are separable,
using BLAS matrix-matrix multiplication~\cite{cct}, or via a Fast Fourier Transform (FFT)~\cite{fftlecun2014}. 

The cost model for the matrix-vector product scheme takes $O(dbk(n-k+1)^2 + bk^3)$ time, but only works when
filters are linearly separable. Meanwhile, the matrix-matrix multiplication scheme has a cost of $O(dbk^2(n-k+1)^2)$. 
Finally, the FFT based scheme takes $O(6dbn^2\log{n}+4dbn^2)$, and the time taken does not depend on k. 

\begin{figure}[!t]
  \centering
  \includegraphics[width=0.8\columnwidth]{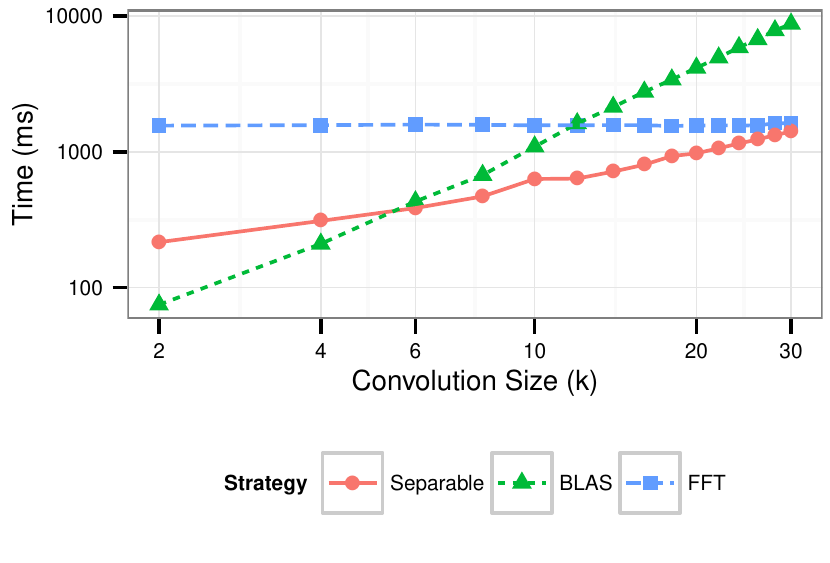}
  \vspace{-0.1in}
  \caption{Time to perform 50 convolutions on a 256x256 3-channel image. As convolution size increases, the optimal method changes.}
  \label{fig:convcomparison}
  \vspace{-0.2in}
\end{figure}

To illustrate the tradeoffs between these methods, in Figure~\ref{fig:convcomparison}, we vary the
size of the convolution filter, $k$, and use representative input images and batch sizes. 
For small values of $k$, we see that BLAS the is fastest operator. However, as $k$ grows, the
algorithm's dependence on $k^2$ makes this approach inappropriate. If the filters are separable, it
is faster to use the matrix-vector algorithm. The FFT algorithm does not depend on $k$ and
thus performs the same regardless of $k$.

\noindent\textbf{Cost Model Evaluation:} To evaluate how well our cost-model works, we compared the
physical operator chosen by our optimizer against the best choice from empirically 
measured values for linear solvers (Figure~\ref{fig:solvercomparisontime}) and PCA (Table~\ref{tab:pcacomparison}).
We found that our optimizer made the right choice 90\% of the time for linear solvers and 84\% of
the time for PCA. In both cases we found that the wrong choices were made when the running time 
of two operators were close to each other and thus the approximate cost model did not severely impact 
overall performance. For example, for the linear solver with 4096 dense features, the 
optimizer chooses the BlockSolver but empirically the Exact solver is about 30\%
faster.

As seen from the three examples above, the choice of optimal physical execution depends on
hardware properties and on properties of the input data. Thus, choices made in support of operator-level optimization depend on upstream processing
and cheaply estimating data properties at various points in the pipeline is an important problem.  
We next discuss how operator chaining semantics can help in achieving this.


%% file: pipelineopt.tex
\section{Whole-Pipeline Optimization}
\label{sec:pipelineopt}

\subsection{Execution Subsampling}
Operator optimization in {\keystone} requires the collection of statistics about input data at each pipeline stage.
For example, a text featurization operator might map a string into a $10,000$-dimensional sparse feature vector. Without statistics about the input (e.g. vector sparsity) after featurization, a downstream operator will be unable to make its optimization decision.
As such, dataset statistics ($A_s$) are determined by first estimating the size of the initial input dataset (in records), and optimizing the first operator in the pipeline with statistics derived from a sample of the input data.
The optimized operator is then executed on the sample, and subsequent operators are optimized.
This procedure continues until all nodes have been optimized.
Along the way, we form a \emph{pipeline profile}, which includes not just the information needed to form $A_s$ at each step, but also information about operator execution time and memory consumption of each operator's execution on the sample. 
We use the pipeline profile to inform the Automatic Materialization optimization described below. We also evaluate the overheads from profiling in Section~\ref{subsec:eval-opt}.

\subsection{Common Sub-expression Elimination}
\label{sec:subexprelim}
One of the whole-pipeline rewrites done by {\keystone} is a form of common sub-expression elimination. 
It is common for training data or the output of featurization stages to be used in several stages of a pipeline. 
As a concrete example, in a text classification pipeline we might first tokenize the training data then determine the $100,000$ most common bigrams in a text corpus, featurize the data to a binary vector indicating the presence of each bigram, and then train a classifier on the same training data. 
Thus, we need the bigrams of each document \emph{both} in the most common features calculation as well as when training the classifier.
{\keystone} identifies and merges such common sub-expressions to enable computation reuse.

\subsection{Automatic Materialization}
Cache management and automatic selection of materialized views are important optimizations used by database management systems~\cite{chirkova} and they have been studied in the context of analytical query systems~\cite{zilio04,ullman96}, and feature selection~\cite{zhang2014feature}. For ML workloads, materialization of intermediate data is very important for performance because the iterative nature of these workloads means that recomputation costs are multiplied across iterations.  
By capturing the iterative nature of the pipelines in the DAG, our optimizer is capable of identifying opportunities for reuse, eliminating redundant computation.
We next describe a formulation for the materialization problem in iterative pipelines and propose an algorithm to automatically select a good set of intermediate objects to materialize in order to speed up ML pipeline execution.

Given the depth-first execution model and the deterministic and side-effect free nature of {\keystone} operators, a natural strategy is materialization of operator outputs that are visited multiple times during the execution. 
This optimization works well in the absence of memory constraints.
 
However, in many applications we have built with {\keystone}, intermediate output can grow to multiple terabytes in size, even for modestly sized inputs. 
On current hardware, this output is too big to fit in memory, even with hundreds of GB of memory per machine. 
Commonly used caching policies such as LRU can result in suboptimal run times because the decision to cache a large object (e.g. intermediate features) may evict a smaller object that is needed later in the pipeline and may be expensive to recompute (e.g. image features). 

We propose an algorithm to automatically select the items to cache in the presence of memory constraints, given that we know how often the objects will be accessed, that we can estimate their size, and that we can estimate the runtime associated with materializing them.

\begin{algorithm}[!tbh]
  \small
 \SetKwInOut{Input}{input}\SetKwInOut{Output}{output} 
 \SetKwFunction{GreedyOptimizer}{GreedyOptimizer}\SetKwFunction{pickNext}{pickNext}\SetKwFunction{stillRoom}{stillRoom}\SetKwFunction{estRuntime}{estRuntime}
 \SetKwBlock{Algorithm}{Algorithm \GreedyOptimizer:}{end}
 \Algorithm{
     \Input{G, t, size, memSize}
     \Output{cache}
     cache $\gets$ $\emptyset$\;
     memLeft $\gets$ memSize\;
     
     next $\gets$ \pickNext(G, cache, size, memLeft, t)\;
     
     \While{nextNode $\ne$ $\emptyset$}{
        cache $\gets$ cache $\cup$ next\;
        memLeft $\gets$ memLeft - size(next)\;
        next $\gets$ \pickNext(G, cache, size, memLeft, t)\;
     }
     \Return cache\;
 }
 \setcounter{AlgoLine}{0}
 \SetKwBlock{Procedure}{Procedure \pickNext:}{end}
 \Procedure{
    \Input{G, cache, size, memLeft, t}
    \Output{next}
    minTime $\gets$ $\infty$\;
    next $\gets$ $\emptyset$\;
    
    \For{v $\in$ nodes(G)}{
        runtime $\gets$ \estRuntime(G, cache $\cup$ v, t)\;
        \If{runtime $<$ minTime $\&$ size(v) $<$ memLeft} {
            next $\gets$ v\;
            minTime $\gets$ runtime\;
        }
    }
    \Return next\;
 }
 \caption{The caching algorithm in {\keystone} builds a cache set by finding the node that will maximize time saved subject to memory constraints. \texttt{estRuntime} is a procedure that computes $T(v)$ for a given DAG, cache set, and node.}
 \label{alg:greedycache}
\end{algorithm}

We formulate the problem as follows: Given a memory budget, we want to find the set of nodes to include in the cache set that minimizes total execution time.

Let $v$ be our node of interest in a pipeline $G$, $t(v)$ is the time taken to do the computation that is local to node $v$ per iteration, $C(v)$ is the number of times a node will by called by its direct successors during execution, and $w_v$ is the number of times a node iterates over its inputs. $T(n)$, the total execution time of the pipeline up to and including node $v$ is:

\[ T(v) = \frac{w_v(t(v) + \sum\limits_{c \in \chi(v)} T(c))}{C(v)^{\kappa_v}} \]

\noindent where $\kappa_v \in \{0,1\}$ is a binary indicator variable signifying whether a node is cached or not, and $\chi(v)$ represents the direct predecessors of $v$ in the DAG.

Where $C(v)$ is defined as follows:
\[ C(v) = \begin{cases}
        \sum\limits_{p \in \pi(v)} w_p C(p)^{\kappa_p},& |\pi(v)| > 0 \\ 
        1, &\text{otherwise}
        \end{cases}
\]

\noindent where $\pi(v)$ represents the direct successors of $v$ in the DAG. 
Because of the DAG structure of the pipeline graph, we are guaranteed to not have any cycles in this graph, thus both $T(v)$ and $C(v)$ are well-defined. 

We can state the problem of minimizing pipeline execution time formally as an optimization problem with linear constraints as follows:
\[ \min\limits_{\kappa} T(sink(G)) \]
\[ s.t. \sum_{v \in V} size(v)\kappa_v \le memSize \]

Where $sink(G)$ is the pipeline terminus, $size(v)$ the size of $v$'s output, and $memSize$ the memory constraint. 

This problem can also be thought of as problem of finding an optimal cache schedule. It is tempting to reach for classical results~\cite{belady1966study,raghavan1989memory} in the optimal paging literature to identify an optimal or near-optimal schedule for this problem. However, neither of these results matches our problem setting fully. In particular, Belady's algorithm is only optimal when each item has a fixed cost to bring into cache (as is common in reads from a two-level memory hierarchy), while in our problem these costs are variable and depend heavily on the computation time to materialize them--in many cases recomputing may be two orders of magnitude faster than reading from disk but an order of magnitude slower than reading from memory, and each operator will have a different computational profile. Second, algorithms for the weighted paging problem don't take into account weights that are dependent on the current state of the cache. e.g. it may be much faster to compute image features if images are already in cluster memory than if they need to be retrieved from disk.

However, it is possible to rewrite the optimization problem above as a mixed-integer linear program (ILP), but in our experiments the cost of solving these problems for reasonably complex pipelines with high end ILP solvers was prohibitive for practical use~\cite{gurobi} at optimization time.
Instead, we implement the greedy Algorithm~\ref{alg:greedycache}. Given an unoptimized pipeline DAG, the algorithm chooses to cache the node which will lead to the largest savings in terms of execution time but whose output fits in available memory.
This process proceeds iteratively until either no benefit to additional caching is possible or all available memory has been used.

%% file: evaluation.tex
\begin{table*} \centering
  \footnotesize
\begin{tabular}{ccccccccc}
\\[-1.8ex]\hline
\hline \\[-1.8ex] 
Dataset & Train Size & Num Train & Test Size & Num Test & Classes & Type & Solve Features & Solve Size \\
        & (GB) &  & (GB)   &          &         &      &                &  (GB) \\
\hline \\[-1.8ex]
Amazon & 13.97 & 65000000 & 3.88 & 18091702 & 2 & text & 100000 (0.1\% sparse) & 89.1 \\
TIMIT & 7.5 & 2251569 & 0.39 & 115934 & 147 & 440-dim vector & 528000 (dense) & 8857 \\
ImageNet & 74 & 1281167 & 3.3 & 50000 & 1000 & 10k pixels image & 262144 (dense) & 2502 \\
VOC & 0.428 & 5000 & 0.420 & 5000 & 20 & 260k pixels image & 40960 (dense) & 1.52 \\
CIFAR-10 & 0.500 & 500000 & 0.001 & 10000 & 10 & 1024 pixels image & 135168 (dense) & 62.9 \\
Youtube8m & 22.07 & 5786881 & 6.3 & 1652167 & 4800 & 1024-dim vector & 1024 (dense) & 44.15 \\
\hline \\[-1.8ex]
\end{tabular}
  \caption{Dataset Characteristics. While raw input sizes may be modest, intermediate state may grow by orders of magnitude before being input to a solver.}
  \label{tab:datasets}
  \vspace{-0.15in}
\end{table*}

\section{Evaluation}
\label{sec:evaluation}

To evaluate the effectiveness of {\keystone}, we explore its ability to efficiently support large scale ML applications in three domains. We also compare {\keystone} with other systems for large scale ML and show how our high-level operators and optimizations can improve performance.
Following that we break down the end-to-end benefits of the previously discussed optimizations.
Finally, we assess the system's ability to scale and show that {\keystone} scales well by enabling the development of scalable, composable components.

\noindent\textbf{Implementation:}  We implement {\keystone} on top of Apache Spark, a cluster computing engine that has been shown to have good scalability and performance for many iterative ML algorithms~\cite{mllib}. 
In {\keystone} we added an additional cache-management layer that is aware of the multiple Spark jobs that comprise a pipeline, and implemented ML operators in the {\keystone} Standard Library that are absent from Spark MLlib.
While the current implementation of the system is Spark-specific, Spark is merely a distributed execution environment and our system can be ported to other backends.

Experiments are run on Amazon EC2 \texttt{r3.4xlarge} instances. 
Each machine has 8 physical cores, 122 GB of memory, and a 320 GB SSD, and was running Apache Spark 1.3.1, Scala 2.10, and HDFS from the CDH4 distribution of 
Hadoop. 
We have also run {\keystone} on Apache Spark 1.5, 1.6 and not encountered any performance regressions.
We use OpenBLAS for numerical operations and Vowpal Wabbit~\cite{langford2007vowpal} v8.0 and SystemML~\cite{ghoting2011systemml} v0.9 in our comparisons.
If not otherwise specified, we run on a 16-node cluster.

\begin{table} \centering
  \footnotesize
\begin{tabular}{ccc}
\\[-1.8ex]\hline
\hline \\[-1.8ex] 
Task & Type & Operators Used  \\
\hline \\[-1.8ex]
Amazon & Text  & LowerCase, Tokenize \\
  Reviews &   & NGrams, TermFrequency \\
  Classification &   & LogisticRegression \\
  \hline \\[-1.8ex]
TIMIT & Speech  & RandomFeatures, Pipeline.gather \\
 Kernel SVM  &    & LinearSolver \\
 \hline \\[-1.8ex]
ImageNet & Image & GrayScale, SIFT, LCS, PCA, GMM \\ 
 Classification  &    & FisherVector, LinearSolver \\
 \hline \\[-1.8ex]
VOC  & Image & GrayScale, SIFT, PCA, GMM \\ 
 Classification    &    & FisherVector, LinearSolver \\
 \hline \\[-1.8ex]
 CIFAR-10 & Image & Windower, PatchExtractor \\
   Classification &     & ZCAWhitener, Convolver, LinearSolver \\
                  &     & SymmetricRectifier, Pooler \\
\hline \\[-1.8ex]
\end{tabular}
\caption{Operators used in constructing pipelines for datasets in Table~\ref{tab:datasets}.}
\label{tab:loc}
\vspace{-0.2in}
\end{table}

\subsection{End-to-End ML Applications}
\label{sec:applications}
To demonstrate the flexibility and generality of the {\keystone} API, we implemented end-to-end machine learning pipelines in several domains including text classification, image classification and speech recognition. We next describe these pipelines and  
compare statistical accuracy and performance results obtained using {\keystone} to previously published results.
We took every effort to recreate these pipelines as they were described by their authors, and made sure that our pipelines achieved comparable or better statistical results than those reported by each benchmark's respective authors. 

The operators used to implement these applications are outlined in Table~\ref{tab:loc}, and the datasets used to train them are described in Table~\ref{tab:datasets}. In each case, the datasets significantly increase in size as part of the featurization process, so at model fitting time the size is substantially larger than the raw data, as shown in the last two columns of the table.
The Solve Size is the size of the dataset that is input to a Linear Solver. This may be too large for available cluster memory, as is the case for TIMIT.
Accuracy results on each dataset achieved with {\keystone} as well as those achieved with the original authors code or (where code was unavailable) as reported in their respective works, are reported in Table~\ref{tab:accuracy}.

\noindent\textbf{Text Analytics}: {\keystone} makes it simple for developers to scale their text pipelines to large datasets. Combined with libraries like CoreNLP~\cite{manning2014corenlp}, {\keystone} allows for scalable implementations of many text classification pipelines such as the one shown in Figure~\ref{lst:amazonlisting}.
We evaluated a text classification pipeline based on~\cite{manning2003optimization} on the Amazon Reviews dataset of 65m product reviews~\cite{mcauley2015inferring} with 100k sparse features. 
We find that {\keystone} matches the statistical performance of a Vowpal Wabbit~\cite{langford2007vowpal} pipeline when run on identical resources with the same solver, finishing in $440s$.

\noindent\textbf{Kernel SVM for Speech Recognition}: Kernel SVMs can be used in many classification scenarios as they can approximate any function. Often their performance has been shown to be much better than simpler generalized linear models~\cite{hsu2003practical}. Kernel evaluations can be efficiently approximated using random feature transformations~\cite{rahimi2007random, sindhwani2014high} and pipelines are a natural way to specify such transformations.
Statistical operators like FFTs and cosine transformations and APIs to merge features help us succinctly describe the pipeline in {\keystone}. 
We evaluated a kernel SVM solver on the TIMIT dataset with 528k features. Using {\keystone} this pipeline runs in 138 minutes on 64 machines. 
By contrast, a 256 node IBM Blue Gene machine with 16 cores per machine takes around 120 minutes~\cite{sindhwani2014high}.
In this case, while {\keystone} may be 11\% slower, it is using only $\frac{1}{8}$ the number of cores to solve this computationally demanding problem.

\noindent\textbf{Image Classification}: Image classification systems are useful in many settings. As images carry local information (i.e. information specific to where in the image a feature appears), locality sensitive techniques, e.g. convolutions or spatially-pooled fisher vectors~\cite{sanchez2013image}, can be used to generate training features. {\keystone} makes it easy to modularize the pipeline to use efficient implementations of image processing operators like SIFT~\cite{lowe1999object} and Fisher Vectors~\cite{sanchez2013image,Chatfield11}. 
Many of the same operators we consider here are necessary components of ``deep-learning'' pipelines~\cite{Krizhevsky2010} which typically train neural networks via stochastic gradient descent and back-propagation.

Using the VOC dataset, we implement the pipeline described in~\cite{Chatfield11}. This pipeline executes end-to-end on 32 nodes using {\keystone} in just 7 minutes. Using the authors original source code the same workload takes 1 hour and 27 minutes to execute on a single 16-core machine with 256 GB of RAM--{\keystone} achieves a $12.4\times$ speedup with $16\times$ the cores. 
We evaluated a Fisher Vector based pipeline on ImageNet with 256k features. The {\keystone} pipeline runs in 4.5 hours on 100 machines. The original pipeline takes four days~\cite{xrce2010imagenet} to run using a highly specialized codebase on a 16-core machine, a $21\times$ speedup on $50\times$ the cores.

In summary, using {\keystone} we achieve one to two orders of magnitude improvement in end-to-end throughput versus a single node, and equivalent or better performance over cluster systems running similar workloads.
These improvements mean much quicker ML application development which leads to higher developer productivity. Next we compare {\keystone} to other large scale learning systems.

\begin{table} \centering
  \footnotesize
\addtolength{\tabcolsep}{-1pt}    

\begin{tabular}{ccccc}
\\[-1.8ex]\hline
\hline \\[-1.8ex] 
Dataset & \multicolumn{2}{c}{\keystone} & \multicolumn{2}{c}{Reported}\\
        & Accuracy &  Time (m) &  Accuracy &  Time (m) \\
\hline \\[-1.8ex]
Amazon~\cite{manning2003optimization} & 91.6\% & 3.3 & - & - \\
TIMIT~\cite{huang2014kernel} & 66.06\% & 138 & 66.33\% & 120 \\
ImageNet~\cite{sanchez2013image}\tablefootnote{We report accuracy on 64k features for ImageNet, while time is reported on 256k features due to lack of consistent reporting by the original authors. The workloads are otherwise similar.} & 67.43\% & 270 & 66.58\% & 5760 \\
VOC 2007~\cite{Chatfield11} & 57.2\% & 7  & 59.2\%   & 87 \\
CIFAR-10~\cite{tensorflowperf} & 84.0\% & 28.7 & 84.0\%   & 50.0 \\
\hline \\[-1.8ex]
\end{tabular}
\addtolength{\tabcolsep}{1pt}
  \caption{Time to Accuracy with {\keystone} obtained on ML pipelines described in the relevant publication. Accuracy for VOC is mean average precision. Accuracy for ImageNet is Top-5 error.}
  \label{tab:accuracy}
  \vspace{-0.15in}
\end{table}

\subsection{{\keystone} vs. Other Systems}
We compare runtimes for the {\keystone} solver with both a specialized system, \textbf{Vowpal Wabbit}~\cite{langford2007vowpal}, built to estimate linear models, and \textbf{SystemML}~\cite{ghoting2011systemml}, a general purpose ML system, which optimizes the implementation of linear algebra operators used in specific algorithms (e.g., Conjugate Gradient Method), but does not choose among logically equivalent algorithms.
We compare solver performance across different feature sizes for two binary classification problems: Amazon and a binary version of TIMIT.
The systems were run with identical inputs and objective functions, and we report end-to-end solve time. For this comparison, we solve binary problems because SystemML does not include a multiclass linear solver.

\begin{figure}
 \includegraphics[width=0.95\columnwidth]{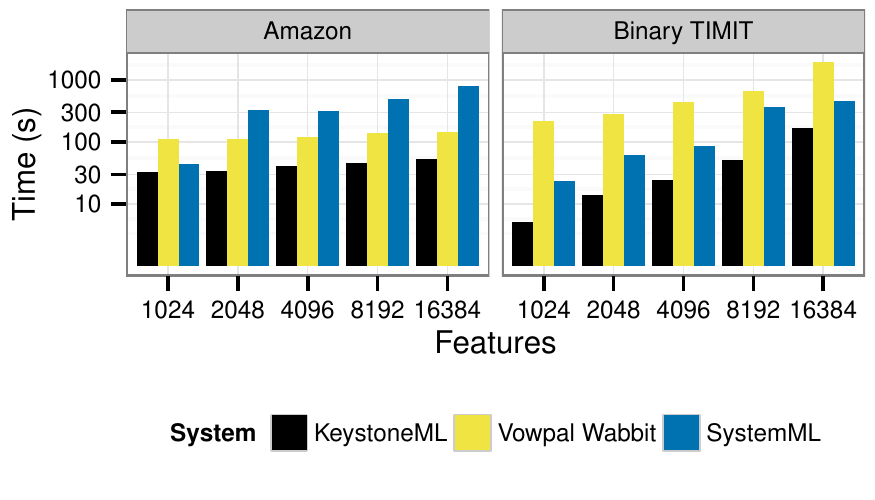}
    \vspace{-0.1in}
    \caption{{\keystone}'s optimizing linear solver outperforms both a specialized and optimizing ML system for two problems across feature sizes. Times are on log scale.}
    \vspace{-0.15in}
    \label{fig:keystonesystems}
    \vspace{-0.1in}
\end{figure}

The results are shown in Figure~\ref{fig:keystonesystems}. The optimized solver in {\keystone} outperforms both Vowpal Wabbit and SystemML because it selects an \emph{appropriate algorithm} to solve the logical problem, as opposed to relying on a one-size fits all operator. At $1024$ features for the Binary TIMIT problem, {\keystone} chooses to run an exact solve, while from $2048$ to $32768$ features it chooses a Dense L-BFGS implementation. At $65536$ features (not pictured), {\keystone} finishes in 17 minutes, while SystemML takes 1 hour and 40 minutes to converge to \emph{worse} training loss over 10 iterations, a speedup of $5.5\times$.

The reasons for these performance differences are twofold: first, since {\keystone} raises the level of abstraction to the logical level, the system can automatically select, for example, a sparse solver for sparse data or an exact algorithm when the number of features is low, or a block solver when the features are high. In the middle, particularly for {\keystone} vs. SystemML on the Binary TIMIT dataset, the algorithms are similar in terms of complexity and access patterns. In this case {\keystone} is faster because feature extraction is \emph{pipelined} with the solver, while SystemML requires a conversion process for data to be fed into a format suitable for the solver. If we \emph{only} consider the solve step of the pipeline, {\keystone} is roughly $1.5\times$ faster than SystemML for this problem.

\textbf{TensorFlow} is a newly open-sourced ML system developed by Google~\cite{tensorflow2015-whitepaper}.
Developed concurrently to {\keystone}, TensorFlow also represents pipelines as graph of dataflow operators.
However, the design goals of the two systems are fundamentally different.
{\keystone} is designed to support horizontally scalable workloads to offer good scale out performance for conventional machine learning applications consisting of featurization and model estimation, while TensorFlow is designed to support neural network models trained via mini-batch SGD with back-propagation. We compare against TensorFlow v0.8 and adapt a multi-GPU example~\cite{tensorflowperf} to a distributed setting in a procedure similar to~\cite{tensorflowsgd}.

To illustrate the differences, we compare the systems' performance on CIFAR-10 top-1 classification performance.
While the learning tasks are identical (i.e., make good predictions on a test dataset, given a training dataset), the workloads are not identical. Specifically, TensorFlow implements a model similar to the one presented in~\cite{Krizhevsky2010}, while in {\keystone} we implement a version of the model similar to~\cite{coates2012learning}.
TensorFlow was run with default parameters and we experimented with strong scaling (fixed 128 image batch size) and weak scaling (batch size of $128 \times Machines$).

\begin{table}[!t]
\footnotesize
\centering
\begin{tabular}{lllllll}
\\[-1.8ex]\hline
\hline \\[-1.8ex]
Machines            & 1     & 2     & 4     & 8     & 16    & 32    \\
\hline \\[-1.8ex]
TensorFlow (strong) & 184 & 90 & 57 & 67  & 122 & 292 \\
TensorFlow (weak)   & 184 & 135 & 135 & 114 & xxx & xxx \\
KeystoneML          & 235 & 125 & 69  & 43  & 32  & 29 \\
\hline \\[-1.8ex]
\end{tabular}
\caption{Time, in minutes, to 84\% accuracy on the CIFAR-10 dataset with {\keystone} and TensorFlow configured for both strong and weak scaling. In large weak scaling regimes TensorFlow failed to converge to a good model.}
\label{tab:keystone-tensorflow-scaling}
\vspace{-0.1in}
\end{table}

For this workload, TensorFlow achieves its best performance on 4-node cluster with 32 total CPU cores, running in 57 minutes. Meanwhile, {\keystone} surpasses its performance at 8 nodes and continues to improve in total runtime out to 32 nodes, achieving a minimum runtime of 29 minutes, or a $1.97\times$ speedup. These results are summarized in Table~\ref{tab:keystone-tensorflow-scaling}. We ran TensorFlow on CPUs for the sake of comparability. Prior benchmarks~\cite{tensorflowperf} have shown that the speed of a single multi-core CPU is comparable to a single GPU; thus the same pipeline finishes in 50 minutes on a 4 GPU machine.

TensorFlow's lack of scalability on this task is fundamental to the chosen model and the algorithm being used to fit it. Minimizing a non-convex loss function via minibatch Stochastic Gradient Descent (SGD) requires coordination of the model parameters after a small number of examples are seen. In this case, the coordination requirements surpass the savings from parallelism at a small number of nodes. While TensorFlow has better scalability on some model architectures~\cite{szegedy2015rethinking}, it is not scalable for other architectures. By contrast, by using a communication-avoiding solver we are able to scale out {\keystone}'s performance on this task significantly further.

Finally, a recent benchmark dataset from YouTube~\cite{youtube8m} describes learning pipelines involving featurization with a neural network~\cite{szegedy2015rethinking} followed by a logistic regression model or SVM. The authors claim that ``models train to convergence in less than a day on a single machine using the publicly-available TensorFlow framework.'' We performed a best-effort replication of this pipeline using {\keystone}. We are unable to replicate the author's claimed accuracy--our pipeline achieves 21\% mAP while they report 28\% mAP. {\keystone} trains a linear classifier on this dataset in 3 minutes, and a converged logistic regression model with worse accuracy in 90 minutes (31 batch gradient evaluations) on a 32-node cluster. The ability to choose an appropriate solver and readily scale out are the key enablers of {\keystone}'s performance.

We now study the impact of {\keystone}'s optimizations.

\subsection{Optimization Levels}
\label{subsec:eval-opt}
The end-to-end results reported earlier in this section are achieved by taking advantage of the complete set of optimizations available in {\keystone}. To understand how important the per-operator and whole-pipeline optimizations described in Sections~\ref{sec:nodeopt} and \ref{sec:pipelineopt} are we compare three different levels of optimization: a default unoptimized configuration (\texttt{None}), a configuration where only whole-pipeline optimizations are used (\texttt{Pipe Only}) and a configuration with operator-level and whole-pipeline optimizations (\texttt{{\keystone}}).

\begin{figure}[!t]
  \centering
  \includegraphics[width=0.90\columnwidth]{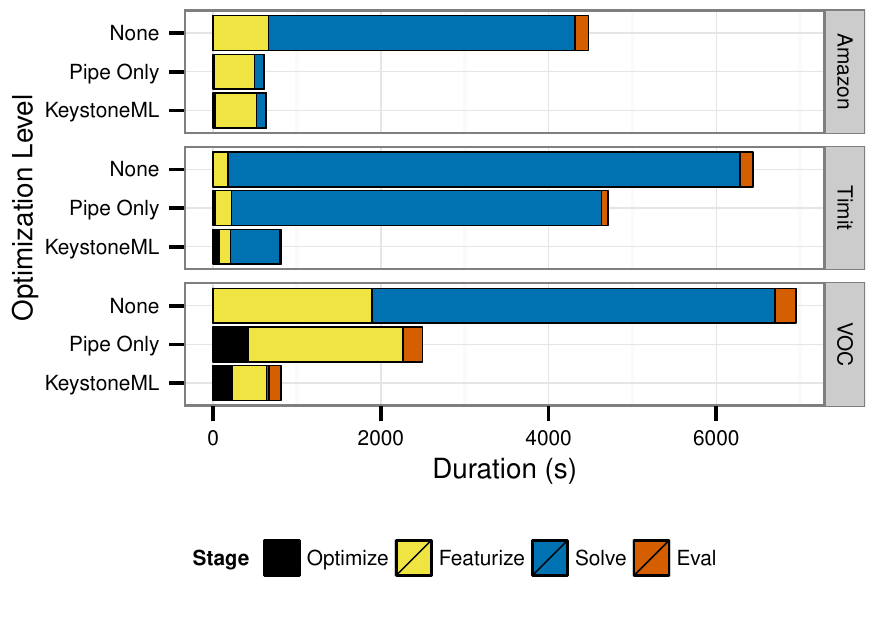}
  \vspace{-0.1in}
  \caption{Impact of optimization levels on three applications, broken down by stage.}
  \label{fig:optimizationlevels}
  \vspace{-0.15in}
\end{figure}

Results comparing these levels, with a breakdown of stage-level timings on the VOC, Amazon and TIMIT pipelines are shown in Figure~\ref{fig:optimizationlevels}. For the Amazon pipeline the whole-pipeline optimizations improve performance by $7\times$, but the operator optimizations do not help further, because the Amazon pipeline uses CoreNLP featurizers which do not have statistical optimizations associated with them, and the default L-BFGS solver turns out to be optimal. The performance gains come from caching intermediate features just before the L-BFGS solve. For the TIMIT pipeline, run with 16k features, we see that the end-to-end optimizations only give a $1.3\times$ speedup but that selecting the appropriate solver results in a $8\times$ speedup over the baseline. 
Finally in the VOC pipeline the whole pipeline optimization gives around $3\times$ speedup. Operator-level optimization chooses good PCA, GMM and solver operators resulting in a $12\times$ improvement over the baseline, or $15\times$ if we amortize the optimization costs across many runs of a similar pipeline. Optimization overheads are insignificant except for the VOC pipeline.
This dataset has relatively few examples, so the sampling strategy takes more time relative to the other datasets.

\begin{figure}[!tbp]
  \centering
  \includegraphics[width=0.9\columnwidth]{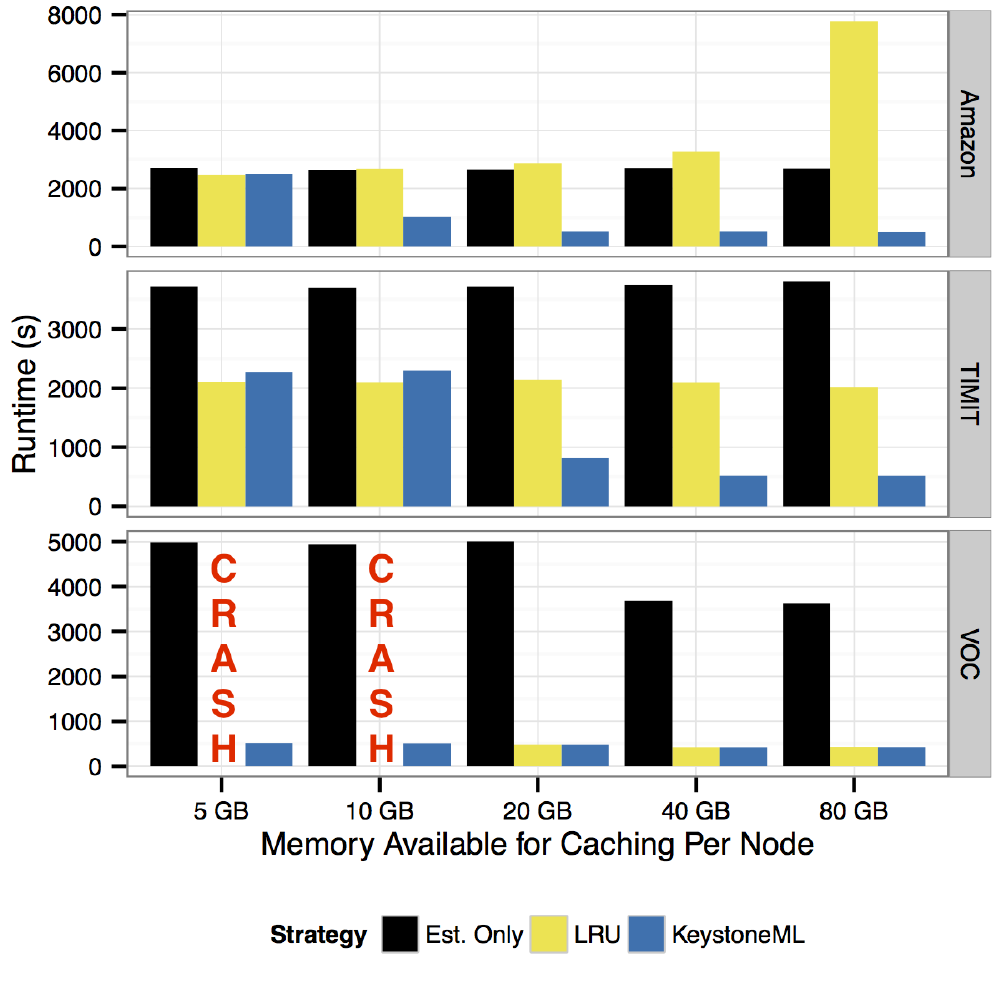}
  \vspace{-0.1in}
  \caption{The {\keystone} caching strategy outperforms a rule-based and LRU caching strategy at many levels of memory constraints and responds well to memory pressure.}
  \label{fig:cachingresults} 
  \vspace{-0.15in}
\end{figure}

\begin{figure}[!t]
  \centering
  \includegraphics[width=1.0\columnwidth]{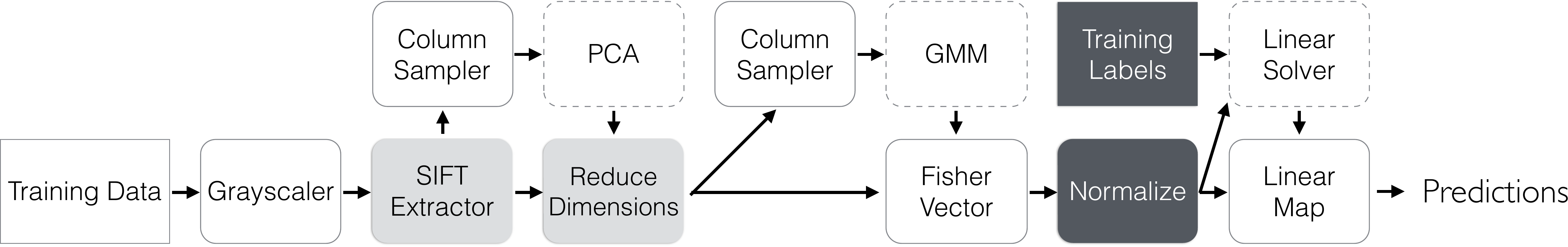}
  \vspace{0.1in}
  \caption{On the VOC workload, the {\keystone} caching strategy selects the colored nodes for caching when allotted 80 GB of cache per machine. As memory resources become scarce, the strategy automatically picks less memory intensive nodes to cache. At smaller scales, only the nodes in dark gray are cached.}
  \label{fig:cachingexample}
  \vspace{-0.1in}
\end{figure}

\subsection{Automatic Materialization Strategies}
\label{subsec:cache-opt}
As discussed in Section~\ref{sec:pipelineopt}, one key optimization enabled by {\keystone}'s ability to capture the complete application DAG to dynamically determine where to materialize reused intermediate objects, particularly in the presence of memory constraints.  In Figure~\ref{fig:cachingresults} we demonstrate the effectiveness of the greedy caching algorithm proposed in Section~\ref{sec:pipelineopt}. Since the algorithm needs local profiles of each node's performance, we measured each node's running time on two samples of 512 and 1024 examples. We extrapolate the node's memory usage and runtime to full scale using linear regression. 
We found that memory estimates from this process are highly accurate and runtime estimates were within $15\%$ of actual runtimes. If estimates are inaccurate, we fall back to an LRU replacement policy for the cache set determined by this procedure. While this measurement process is imperfect, it is adequate at identifying relative running times and thus is sufficient for our purpose of resource management.

We compare this strategy with two alternatives--the first is a simple rule-based approach which only caches the results of Estimators. This is a sensible rule to follow, as the result of an Estimator (a Transformer or model) is computationally expensive to acquire and typically holds a small memory footprint. However, this is not sufficient for most practical pipelines because if a pipeline contains more than one Estimator, often the input to the first Estimator will be used downstream, thus presenting an opportunity for reuse. The second approach is a Least Recently Used (LRU) policy: in a regime where memory is unconstrained, LRU matches the ideal strategy and further, LRU is the default memory management strategy used by Apache Spark. However, LRU does not take into account that datasets from other jobs (even ones in the same pipeline) are vying for presence in cluster memory.

From Figure~\ref{fig:cachingresults} we notice several important trends. First, the {\keystone} strategy is nearly always better than either of the other strategies. 
In the unconstrained case, the algorithm is going to remember all reused items as late in their journey through the pipeline as possible. In the constrained case, it will do as least as well as remembering the (small) estimators which are by definition reused later in the pipeline. Additionally, the strategy degrades effectively, mixing between the best performance of the limited-memory rule-based strategy and the LRU based ``cache everything'' strategy which works well in unconstrained settings. 
Curiously, as we increased the memory available to caching per-node, the LRU strategy performed \emph{worse} for the Amazon pipeline. Upon further investigation, this is because Spark has an implicit admission control policy which only allows objects under some proportion of the cache size to be admitted to the cache at runtime. As the cache size gets bigger in the LRU case, massive objects which are not then reused are admitted to the cache and evict smaller objects which \emph{are} reused and thus need to be recomputed. 

To give a concrete example of the optimizer in action, consider the VOC pipeline
(Figure~\ref{fig:cachingexample}). When memory is not unconstrained (80 GB per node), the outputs from the \texttt{SIFT}, \texttt{ReduceDimensions}, \texttt{Normalize} and \texttt{TrainingLabels} are cached. When memory is restricted (5 GB per node) only the output from \texttt{Normalize} and \texttt{TrainingLabels} are cached.

These results show that both per-operator and whole-pipeline optimizations are important for end-to-end performance improvements. 
We next study the scalability of the system on three workloads

\subsection{Scalability}
\label{subsec:scale}
\begin{figure}[!t]
  \centering
  \includegraphics[width=0.95\columnwidth]{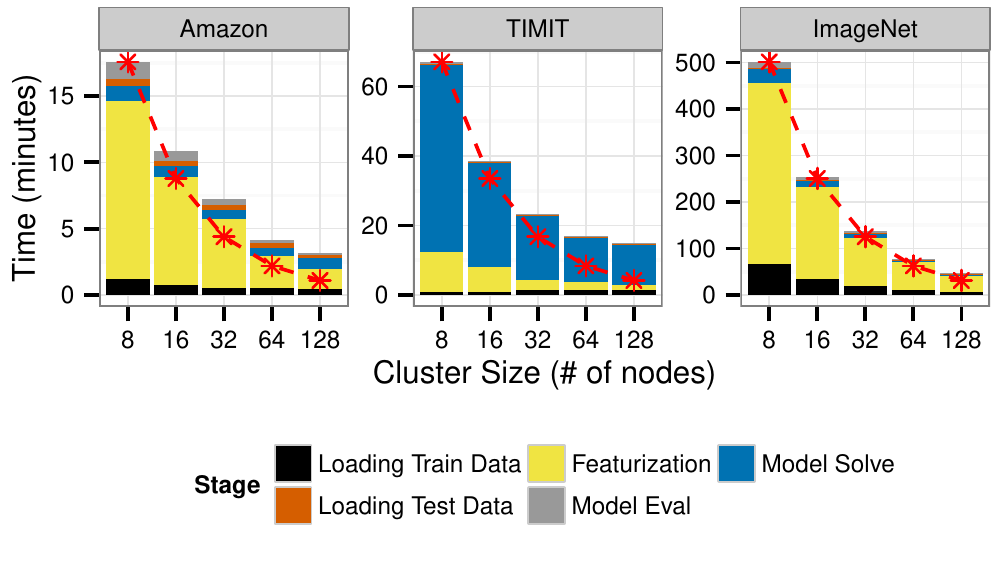}
  \vspace{-0.1in}
  \caption{Time breakdown of workloads by stage. The red line indicates ideal strong scaling performance over 8 nodes.}
  \label{fig:keystone-scaling-breakdown} 
  \vspace{-0.2in}
\end{figure}

As discussed in previous sections, {\keystone}'s API design encourages the construction of scalable operators. However, some estimators like linear solvers need coordination~\cite{demmel2012communication} among workers to compute correct results.
In Figure~\ref{fig:keystone-scaling-breakdown} we demonstrate the scaling properties from 8 to 128 nodes of the text, image, and Kernel SVM pipelines on the Amazon, ImageNet (with 16k features) and TIMIT datasets (with 65k features) respectively. The ImageNet pipeline exhibits near-perfect horizontal scalability up to 128 nodes, while the Amazon and TIMIT pipeline scale well up to 64 nodes.

To understand why the Amazon and TIMIT pipeline do not scale linearly to 128 nodes, we further analyze the breakdown of time take by each stage. We see that each pipeline is dominated by a different part of its computation. The TIMIT pipeline is dominated by its solve stage, while featurization dominates the Amazon and ImageNet pipelines.
Scaling linear solvers is known to require coordination~\cite{demmel2012communication}, which leads directly to sub-linear scalability of the whole pipeline. Similarly, in the Amazon pipeline, one of the featurization steps uses an aggregation tree which does not scale linearly.

%% file: related.tex
\section{Related Work}
\label{sec:related}

\noindent\textbf{ML Frameworks}: ML researchers have traditionally used MATLAB or R packages to develop ML routines. The importance of feature engineering has led to tools like scikit-learn~\cite{scikit-learn} and KNIME~\cite{berthold2008knime} adding support for featurization for small datasets. Further, existing libraries for large scale ML~\cite{cai2014comparison} like Vowpal Wabbit~\cite{langford2007vowpal}, GraphLab~\cite{low2012distributed}, MLlib~\cite{mllib}, RIOT~\cite{zhang2009riot}, DimmWitted~\cite{zhang2014dimmwitted} focus on efficient implementations of learning algorithms like regression, classification and linear algebra routines. In {\keystone}, we focus on pipelines that include featurization and show how to optimize performance with end-to-end information. Work in Parameter Servers~\cite{li2014parameterserver} has studied how to share model updates. In {\keystone} we implement a high-level API for linear solvers and can leverage parameter servers in our architecture.

Closely related to {\keystone} is SystemML~\cite{ghoting2011systemml} which also uses an optimization based approach to determine the physical execution strategy of ML algorithms. However, SystemML places less emphasis on support for UDFs and featurization, while instead focusing on linear algebra operators which have well specified semantics. To handle featurization we develop an extensible API in {\keystone} which allows for cost profiling of arbitrary nodes and uses these cost estimates to make node-level and whole-pipeline optimizations.
Other work~\cite{zhang2014feature,cct} has looked at optimizing caching strategies and operator selection in the regime of feature selection and feature generation workloads. {\keystone} considers similar problems in the context of distributed ML operators and end-to-end learning pipelines. Developed concurrently to {\keystone} is TensorFlow~\cite{tensorflow2015-whitepaper}. While designed to support different learning workloads the optimizations that are a part of {\keystone} can also be applied to systems like TensorFlow. 

Projects such as Bismarck~\cite{feng2012towards}, MADLib~\cite{hellerstein2012madlib}, and GLADE~\cite{qin2013danac} have proposed techniques to integrate ML algorithms inside database engines. In {\keystone}, we develop a high level API and show how we can achieve similar benefits of modularity and end-to-end optimization while also being scalable. These systems do not present cross-operator optimizations and do not consider tradeoffs at the operator level that we consider in {\keystone}. Finally, Spark ML~\cite{sparkmlpipelines} represents an early design of a similar high-level API for machine learning. We present a type safe API and optimization framework for such a system. 
The version we present in this paper differs in its use of type-safe operations, support for complex
data flows, internal DAG representation and optimizations discussed in
Sections~\ref{sec:nodeopt}~and~\ref{sec:pipelineopt}. Finally, the concept of using a high-level
programming model has been explored in a number of other contexts, including compilers~\cite{llvm}
and networking~\cite{kohler2000click}. In this paper we focus on machine learning workloads and
propose node-level and end-to-end optimizations.

\noindent\textbf{Query Optimization, Modular Design, Caching}:
There are several similarities between the optimizations made by {\keystone} and traditional relational query optimizers. Even the earliest relational query optimizers~\cite{selinger} used multiple physical implementations of equivalent logical operators, and like many relational optimizers, the {\keystone} optimizer is cost-based.  However, {\keystone} supports a much richer set of data types than a traditional relational query system, and our operators lack some relational algebra semantics, such as commutativity, limiting the system's ability to perform certain optimizations. Further, {\keystone} switches among operators that provide exact answers vs approximate ones to save time due to the workload setting. Data characteristics such as sparsity are not traditionally considered by optimizers.

The caching strategy employed by {\keystone} can be viewed as a form of view selection for materialized view maintenance over queries with expensive user-defined functions~\cite{chirkova,hellerstein97}, we focus on materialization for intra-query optimization, as opposed to inter-query optimization~\cite{ullman96,autoadmin,zilio04,restore,perez2014}. While much of the related work focuses on the challenging problem of view maintenance in the presence of updates, {\keystone} we exploit the iterative nature and immutable properties of this state.

%% file: conclusion.tex
\section{Future Work and Conclusion}
{\keystone} represents a significant first step towards easy-to-use, robust, and efficient end-to-end ML at massive scale.
We plan to investigate pipeline optimizations like node reordering to reduce data transfers and also look at how hyperparameter tuning~\cite{sparks2015automating} can be integrated into the system.
The existing {\keystone} operator APIs are synchronous and our existing pipelines are acyclic. In the future we plan to study how algorithms like asynchronous SGD~\cite{li2014parameterserver} or back-propagation can be integrated with the robustness and scalability that {\keystone} provides. 

We have presented the design of {\keystone}, a system that enables the development end-to-end ML pipelines. By capturing the end-to-end application, {\keystone} can automatically optimize execution at both the operator and whole-pipeline levels, enabling solutions that automatically adapt to changes in data, hardware, and other environmental characteristics.